\documentclass{article}

\usepackage[final]{nips_2017}
\usepackage[utf8]{inputenc} 
\usepackage[T1]{fontenc}    
\usepackage{xcolor}         
\usepackage{url}            
\usepackage{booktabs}       
\usepackage{multirow}
\usepackage{amsfonts}       
\usepackage{nicefrac}       
\usepackage{microtype}      
\usepackage[pdftex]{graphicx}
\usepackage{amsmath}    
\usepackage{wrapfig}
\usepackage{subcaption}
\usepackage{array}
\usepackage[hidelinks]{hyperref}

\definecolor{LinkColor}{rgb}{0,0,0}    
\hypersetup{colorlinks=true,
    linkcolor=LinkColor,
    citecolor=LinkColor,
    filecolor=LinkColor,
    menucolor=LinkColor,
    urlcolor=LinkColor}

\newcommand{\forward}{\rightarrow}
\newcommand{\backward}{\leftarrow}
\newcommand{\lstm}[2]{{\rm LSTM} \left( #1 , #2 \right)}
\newcommand{\bilstm}[1]{{\rm biLSTM} \left( #1 \right)}

\newcommand{\dec}{^{\rm dec}}
\newcommand{\softmax}[1]{{\rm softmax} \left( #1 \right)}

\newcommand{\ftanh}[1]{{\rm tanh} \left( #1 \right)}

\title{Learned in Translation: Contextualized Word Vectors}

\author{
Bryan McCann \\
  \texttt{bmccann@salesforce.com}
  \And James Bradbury  \\
  \texttt{james.bradbury@salesforce.com}
  \And Caiming Xiong \\
  \texttt{cxiong@salesforce.com} 
  \And Richard Socher \\
  \texttt{rsocher@salesforce.com}
}

\begin{document}
\maketitle

\begin{abstract}
Computer vision has benefited from initializing multiple deep layers with weights pretrained on large supervised training sets like ImageNet.
Natural language processing (NLP) 
typically sees initialization of only the lowest layer of deep models with pretrained word vectors. 
In this paper, 
we use a deep LSTM encoder from an attentional sequence-to-sequence model trained for machine translation (MT) to contextualize word vectors.
We show that adding these context vectors (CoVe) improves performance over using only unsupervised word and character vectors on a wide variety of common NLP tasks: 
sentiment analysis (SST, IMDb), 
question classification (TREC), 
entailment (SNLI), 
and question answering (SQuAD). 
For fine-grained sentiment analysis and entailment,
CoVe improves performance of our baseline models to the state of the art.
\end{abstract}
\section{Introduction}

Significant gains have been made through transfer and multi-task learning between synergistic tasks.
In many cases, these synergies can be exploited by architectures that rely on similar components. 
In computer vision, convolutional neural networks (CNNs) pretrained on ImageNet~\citep{krizhevsky2012imagenet, Deng2009ImageNetAL} have become the \textit{de facto} initialization for more complex and deeper models. 
This initialization improves accuracy on other related tasks such as visual question answering~\citep{xiong2016dynamic} or image captioning~\citep{lu2016knowing, Socher2014TACL}.

In NLP, distributed representations pretrained with models like Word2Vec~\citep{Mikolov2013b} and GloVe~\citep{Pennington2014} have become common initializations for the word vectors of deep learning models. 
Transferring information from large amounts of unlabeled training data in the form of word vectors has shown to improve performance over random word vector initialization on a variety of downstream tasks, 
e.g. part-of-speech tagging~\citep{Collobert2011}, named entity recognition~\citep{Pennington2014}, and question answering~\citep{Xiong2017};
however, words rarely appear in isolation.
The ability to share a common representation of words in the context of sentences that include them could further improve transfer learning in NLP.

Inspired by the successful transfer of CNNs trained on ImageNet to other tasks in computer vision,
we focus on training an encoder for a large NLP task and transferring that encoder to other tasks in NLP.
Machine translation (MT) requires a model to encode words in context so as to decode them into another language,
and attentional sequence-to-sequence models for MT often contain an LSTM-based encoder, 
which is a common component in other NLP models.
We hypothesize that MT data in general holds potential
comparable to that of ImageNet 
as a cornerstone for reusable models.
This makes an MT-LSTM pairing in NLP a natural candidate for mirroring the ImageNet-CNN pairing of computer vision.

\begin{figure}
  \centering
 \includegraphics[width=\textwidth]{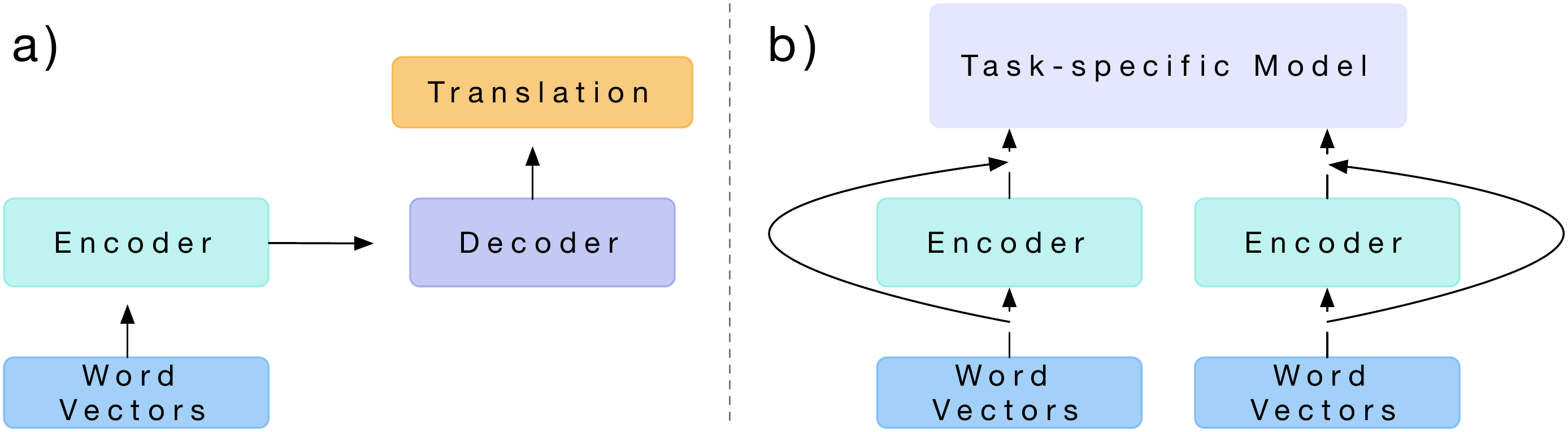}
  \caption{We a) train a two-layer, bidirectional LSTM as the encoder of an attentional sequence-to-sequence model for machine translation and b) use it to provide context for other NLP models.
  }\label{fig1}
  \vspace{-0.3cm}
\end{figure}

As depicted in Figure~\ref{fig1}, 
we begin by training LSTM encoders on several machine translation datasets, 
and we show that these encoders can be used to improve performance of models trained for other tasks in NLP.
In order to test the transferability of these encoders, 
we develop a common architecture for a variety of classification tasks,
and we modify the Dynamic Coattention Network for question answering~\citep{Xiong2017}.
We append the outputs of the MT-LSTMs, 
which we call context vectors (CoVe), 
to the word vectors typically used as inputs to these models.
This approach improved the performance of models for downstream tasks over that of baseline models using pretrained word vectors alone.
For the Stanford Sentiment Treebank (SST) and the Stanford Natural Language Inference Corpus (SNLI),
CoVe pushes performance of our baseline model to the state of the art.

Experiments reveal that the quantity of training data used to train the MT-LSTM is positively correlated with performance on downstream tasks. 
This is yet another advantage of relying on MT, 
as data for MT is more abundant than for most other supervised NLP tasks, 
and it suggests that higher quality MT-LSTMs carry over more useful information.
This reinforces the idea that machine translation is a good candidate task for further research into models that possess a stronger sense of natural language understanding.

\section{Related Work}

\textbf{Transfer Learning.}
Transfer learning, or domain adaptation, has been applied in a variety of areas where researchers identified synergistic relationships between independently collected datasets.~\citet{Saenko2010AdaptingVC} 
adapt object recognition models developed for one visual domain to new imaging conditions by learning a transformation that minimizes domain-induced changes in the feature distribution.~\citet{Zhu2011HeterogeneousTL}
use matrix factorization to incorporate textual information into tagged images to enhance image classification.
In natural language processing (NLP),~\citet{Collobert2011} leverage representations learned from unsupervised learning to improve performance on supervised tasks like named entity recognition, part-of-speech tagging, and chunking.
Recent work in NLP has continued in this direction by using pretrained word representations to improve models for
entailment~\citep{Bowman2014},
sentiment analysis~\citep{Socher2013EMNLP},
summarization~\citep{Nallapati2016AbstractiveTS},
and question answering~\citep{Seo2017BidirectionalAF,Xiong2017}.~\citet{Ramachandran2016UnsupervisedPF} propose initializing sequence-to-sequence models with pretrained language models and fine-tuning for a specific task.~\citet{Kiros2015SkipThoughtV} propose an unsupervised method for training an encoder that outputs sentence vectors that are predictive of surrounding sentences.
We also propose a method of transferring higher-level representations than word vectors, but we use a supervised method to train our sentence encoder and show that it improves models for text classification and question answering without fine-tuning.

\textbf{Neural Machine Translation.} 
Our source domain of transfer learning is machine translation, 
a task that has seen marked improvements in recent years with the advance of neural machine translation (NMT) models.~\citet{Sutskever2014} investigate sequence-to-sequence models that consist of a neural network encoder and decoder for machine translation.~\citet{Bahdanau2015} propose the augmenting sequence to sequence models with an attention mechanism that gives the decoder access to the encoder representations of the input sequence at each step of sequence generation.~\citet{Luong2015EffectiveAT} further study the effectiveness of various attention mechanisms with respect to machine translation.
Attention mechanisms have also been successfully applied to NLP tasks like
entailment~\citep{conneau2017supervised},
summarization~\citep{Nallapati2016AbstractiveTS},
question answering~\citep{Seo2017BidirectionalAF,Xiong2017,min2017question},
and semantic parsing~\citep{Dong2016LanguageTL}.
We show that attentional encoders trained for NMT transfer well to other NLP tasks.

\textbf{Transfer Learning and Machine Translation.}
Machine translation is a suitable source domain for transfer learning because the task,
by nature, 
requires the model to faithfully reproduce a sentence in the target language without losing information in the source language sentence.
Moreover, there is an abundance of machine translation data that can be used for transfer learning.~\citet{Hill2016LearningDR} study the effect of transferring from a variety of source domains to the semantic similarity tasks in~\citet{Agirre2014SemEval2014T1}.~\citet{Hill2017} further demonstrate that fixed-length representations obtained from NMT encoders outperform those obtained from monolingual (e.g. language modeling) encoders on semantic similarity tasks.
Unlike previous work, 
we do not transfer from fixed length representations produced by NMT encoders. 
Instead, we transfer representations for each token in the input sequence.
Our approach makes the transfer of the trained encoder more directly compatible with subsequent LSTMs, attention mechanisms, and, in general, layers that expect input sequences.
This additionally facilitates the transfer of sequential dependencies between encoder states.

\textbf{Transfer Learning in Computer Vision.} 
Since the success of CNNs on the ImageNet challenge, 
a number of approaches to computer vision tasks have relied on pretrained CNNs as off-the-shelf feature extractors.~\citet{girshick2014rich} show that using a pretrained CNN to extract features from region proposals improves object detection and semantic segmentation models.~\citet{qi2016hedged} propose a CNN-based object tracking framework, 
which uses hierarchical features from a pretrained CNN (VGG-19 by~\citet{simonyan2014very}).
For image captioning,~\citet{lu2016knowing} train a visual sentinel with a pretrained CNN and fine-tune the model with a smaller learning rate.
For VQA,~\citet{fukui2016multimodal} propose to combine text representations with visual representations extracted by a pretrained residual network~\citep{he2016deep}.
Although model transfer has seen widespread success in computer vision, 
transfer learning beyond pretrained word vectors is far less pervasive in NLP.

\section{Machine Translation Model}

We begin by training an attentional sequence-to-sequence model for English-to-German translation based on~\citet{2017opennmt} with the goal of transferring the encoder to other tasks.

For training, 
we are given a sequence of words in the source language 
$w^{x}=[w^{x}_1, \ldots, w^{x}_n]$ 
and a sequence of words in the target language
$w^{z}=[w^{z}_1, \ldots, w^{z}_m]$. 
Let GloVe($w^x$)
be a sequence of GloVe vectors corresponding to the words in $w^x$,
and let $z$ be a sequence of randomly initialized word vectors corresponding to the words in $w^z$.

We feed GloVe($w^x$) to a standard, two-layer, bidirectional, long short-term memory network
\footnote{
Since there are several biLSTM variants, we define ours as follows. Let $h=[h_1, \ldots, h_n]=\bilstm{x}$ represent the output sequence of our biLSTM operating on an input sequence $x$. Then a forward LSTM computes $h^{\forward}_t = \lstm{x_t}{h^{\forward}_{t-1}}$ 
for each time step, and a backward LSTM computes 
$h^{\backward}_t = \lstm{x_t}{h^{\backward}_{t+1}}$.
The final outputs of the biLSTM for each time step are 
$h_t = \left[ h^{\forward}_t; h^{\backward}_t \right]$.
}~\citep{graves2005framewise} 
that we refer to as an MT-LSTM 
to indicate that it is this same two-layer BiLSTM that we later transfer as a pretrained encoder.
The MT-LSTM  is used to compute a sequence of hidden states
\begin{equation}
h = \text{\rm MT-LSTM}(\text{GloVe}(w^x)).
\label{eq:mtlstm}
\end{equation}

For machine translation, 
the MT-LSTM supplies the context for an attentional decoder 
that produces a distribution over output words 
$p(\hat{w}^z_t|H,w^z_1,\ldots,w^z_{t-1})$ at each time-step.

At time-step $t$, 
the decoder first uses a two-layer, 
unidirectional LSTM to produce a hidden state $h\dec_t$ 
based on the previous target embedding $z_{t-1}$ 
and a context-adjusted hidden state $\tilde h_{t-1}$:
\begin{equation}
h\dec_{t} = \lstm{ [ z_{t-1}; \tilde h_{t-1} ] }{h\dec_{t-1}}.
\end{equation}

The decoder then computes a vector of attention weights $\alpha$ 
representing the relevance of each encoding time-step to the current decoder state.
\begin{equation}
\alpha_{t} = \softmax{H (W_1 h\dec_{t} + b_1)}
\end{equation}

where $H$ refers to the elements of $h$ stacked along the time dimension.

The decoder then uses these weights as coefficients in an attentional sum
that is concatenated with the decoder state 
and passed through a $\rm tanh$ layer
to form the context-adjusted hidden state $\tilde{h}$:
\begin{equation}
\tilde h_{t} = \ftanh{W_2 \left[ H^\top \alpha_t ; h\dec_{t} \right ] + b_2}
\end{equation}

The distribution over output words is generated by a final transformation of the context-adjusted hidden state:
$
p(\hat{w}^z_t|X,w^z_1,\ldots,w^z_{t-1}) = \softmax{W_{\rm out} \tilde h_t + b_{\rm out}}.
$
\section{Context Vectors (CoVe)}

We transfer what is learned by the MT-LSTM to downstream tasks by treating the outputs of the MT-LSTM as context vectors. 
If $w$ is a sequence of words and GloVe$(w)$ the corresponding sequence of word vectors produced by the GloVe model,
then
\begin{equation}
\text{CoVe}(w) = \text{MT-LSTM}(\text{GloVe}(w))
\label{eq:contextVectors}
\end{equation}
is the sequence of context vectors produced by the MT-LSTM. 
For classification and question answering, 
for an input sequence $w$, 
we concatenate each vector in GloVe($w$) with its corresponding vector in CoVe($w$)
\begin{equation}
\tilde w = [\text{GloVe}(w); \text{CoVe}(w)]
\label{eq:concatVectors}
\end{equation}
as depicted in Figure~\ref{fig1}b.
\section{Classification with CoVe}

\begin{wrapfigure}[30]{r}{0.45\textwidth}
  \centering
  \includegraphics[width=0.4\textwidth]{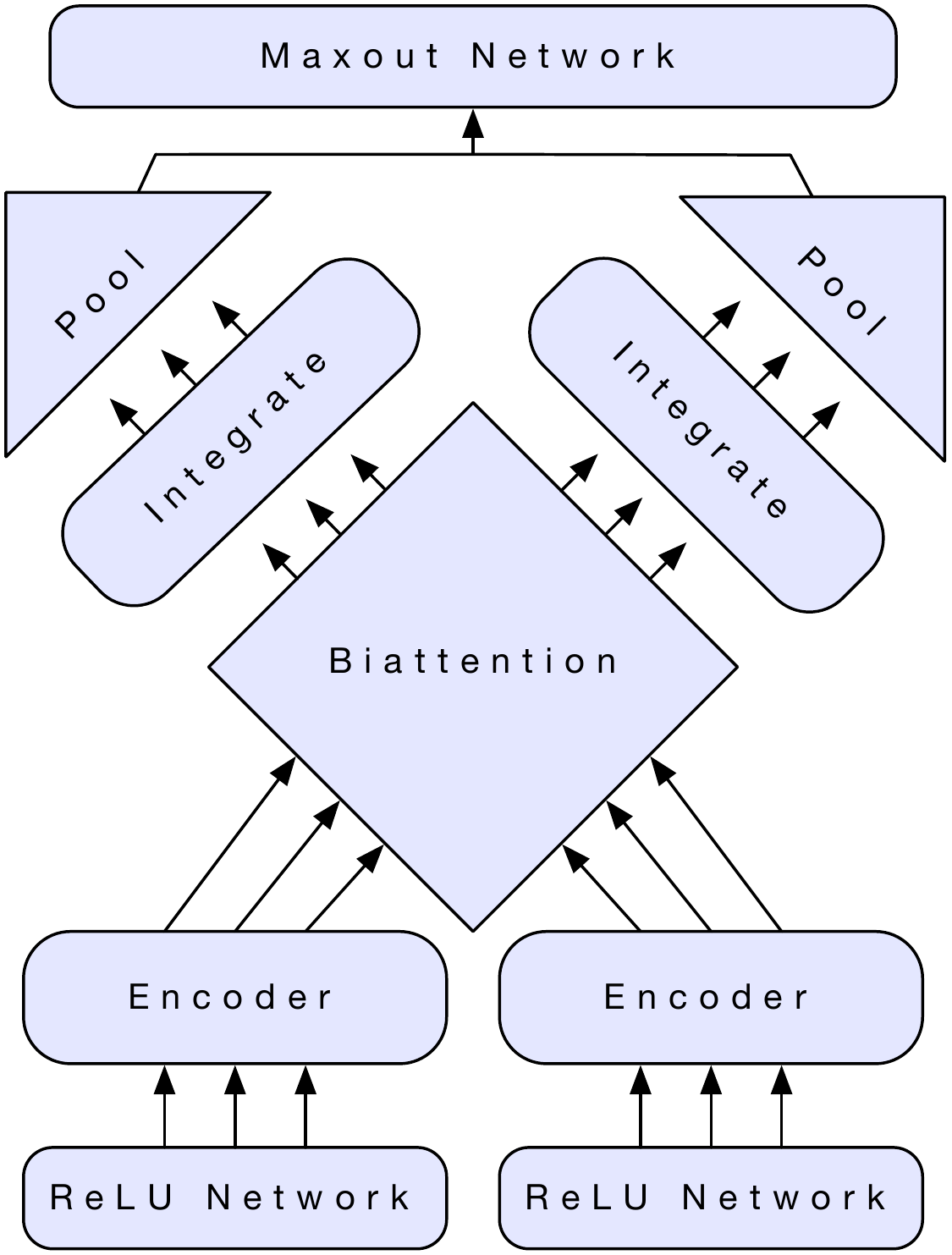}
  \caption{Our BCN uses a feedforward network with ReLU activation and biLSTM encoder to create task-specific representations of each input sequence. Biattention conditions each representation on the other, a biLSTM integrates the conditional information, and a maxout network uses pooled features to compute a distribution over possible classes.
  }
  \label{classarch}
\end{wrapfigure}

We now describe a general biattentive classification network (BCN) we use to test how well CoVe transfer to other tasks.
This model, 
shown in Figure~\ref{classarch},
is designed to handle both single-sentence and two-sentence classification tasks.
In the case of single-sentence tasks, 
the input sequence is duplicated to form two sequences,
so we will assume two input sequences for the rest of this section.

Input sequences $w^x$ and $w^y$ are converted to sequences of vectors, $\tilde w^x$ and $\tilde w^y$, as described in  Eq.~\ref{eq:concatVectors} before being fed to the task-specific portion of the model (Figure~\ref{fig1}b).

A function $f$ applies a feedforward network with ReLU activation~\citep{Nair10} 
to each element of  $\tilde w^x$ and $\tilde w^y$, 
and a bidirectional LSTM processes
the resulting sequences to obtain task specific representations, 
\begin{equation}
x = \bilstm{f(\tilde w^x)}
\label{eq:taskVectorsX}
\end{equation}
\begin{equation}
y = \bilstm{f(\tilde w^y)}
\label{eq:taskVectorsY}
\end{equation}
These sequences are each stacked along the time axis to get matrices $X$ and $Y$.

In order to compute representations that are interdependent, 
we use a biattention mechanism~\citep{Seo2017BidirectionalAF,Xiong2017}.
The biattention first computes an affinity matrix $A=XY^\top$. 
It then extracts attention weights with column-wise normalization:
\begin{equation}
A_x=\softmax{A}
\qquad
A_y=\softmax{A^\top}
\end{equation}
which amounts to a novel form of self-attention when $x=y$. 
Next, 
it uses context summaries 
\begin{equation}
C_x=A^\top_x X
\qquad 
C_y=A_y^\top Y
\end{equation}
to condition each sequence on the other.\bigskip

We integrate the conditioning information into our representations
for each sequence with two separate one-layer, 
bidirectional LSTMs that operate on the concatenation of
the original representations
(to ensure no information is lost in conditioning), 
their differences from the context summaries
(to explicitly capture the difference from the original signals),
and the element-wise products between originals and context summaries
(to amplify or dampen the original signals).
\begin{align}
X_{|y} &= \bilstm{ \left[X; X - C_y; X \odot C_y \right]}
\\
Y_{|x} &= \bilstm{ \left[Y; Y - C_x; Y \odot C_x \right] }
\end{align}

The outputs of the bidirectional LSTMs are aggregated by pooling along the time dimension.
Max and mean pooling have been used in other models to extract features, 
but we have found that adding both min pooling 
and self-attentive pooling can aid in some tasks.
Each captures a different perspective on the conditioned sequences.

The self-attentive pooling computes weights for each time step of the sequence 
\begin{equation}
\beta_{x} = \softmax{X_{|y}v_1 + d_1}
\qquad
\beta_{y} = \softmax{Y_{|x}v_2 + d_2}
\end{equation}

and uses these weights to get weighted summations of each sequence:
\begin{equation}
x_{\rm self} = X_{|y}^\top \beta_{x}
\qquad
y_{\rm self} = Y_{|x}^\top \beta_{y}
\end{equation}

The pooled representations are combined to get one joined representation for all inputs. 
\begin{align}
x_{\rm pool} &= \left [ \max(X_{|y}); {\rm mean}(X_{|y}); \min(X_{|y}); x_{\rm self} \right ]
\\
y_{\rm pool} &= \left [ \max(Y_{|x}); {\rm mean}(Y_{|x}); \min(Y_{|x});  y_{\rm self} \right ]
\end{align}

We feed this joined representation through a three-layer, 
batch-normalized~\citep{Ioffe2015BatchNA} maxout network~\citep{Goodfellow2013MaxoutN} 
to produce a probability distribution over possible classes.

\section{Question Answering with CoVe}

For question answering, 
we obtain sequences $x$ and $y$ just as we do in Eq.~\ref{eq:taskVectorsX} and Eq.~\ref{eq:taskVectorsY} for classification, except that the function $f$ is replaced with a function $g$ that uses a tanh activation instead of a ReLU activation.
In this case, one of the sequences is the document and the other the question in the question-document pair.
These sequences are then fed through the coattention and dynamic decoder implemented as in the original Dynamic Coattention Network (DCN) ~\citep{xiong2016dynamic}. 
\section{Datasets}

\textbf{Machine Translation.} 
We use three different English-German machine translation datasets to train three separate MT-LSTMs. 
Each is tokenized using the Moses Toolkit~\citep{Koehn2007}.

Our smallest MT dataset comes from the WMT 2016 multi-modal translation shared task~\citep{Specia2016}.
The training set consists of 30,000 sentence pairs that briefly describe Flickr captions and is often referred to as Multi30k.
Due to the nature of image captions, this dataset contains sentences that are, on average, shorter and simpler than those from larger counterparts.

Our medium-sized MT dataset is the 2016 version of the machine translation task prepared for the International Workshop on Spoken Language Translation~\citep{cettolo2015iwslt}. 
The training set consists of 209,772 sentence pairs from transcribed TED presentations that cover a wide variety of topics with more conversational language than in the other two machine translation datasets.

Our largest MT dataset comes from the news translation shared task from WMT 2017.
The training set consists of roughly 7 million sentence pairs that comes from web crawl data, a news and commentary corpus, European Parliament proceedings, and European Union press releases.

We refer to the three MT datasets as MT-Small, MT-Medium, and MT-Large, respectively,
and we refer to context vectors from encoders trained on each in turn as CoVe-S, CoVe-M, and CoVe-L.
\bigskip

\begin{table}
  \centering
\begin{tabular}{llll}
    \toprule
Dataset & Task & Details & Examples\\
\midrule
 SST-2   &Sentiment Classification & 2 classes, single sentences        & 56.4k\\
 SST-5   & Sentiment Classification & 5 classes, single sentences           &94.2k \\
 IMDb    & Sentiment Classification & 2 classes, multiple sentences          &22.5k   \\
TREC-6  & Question Classification& 6 classes & 5k  \\
  TREC-50 & Question Classification& 50 classes &5k  \\
SNLI    &Entailment Classification &  2 classes        &  550k \\
SQuAD & Question Answering & open-ended (answer-spans)  & 87.6k  \\
\bottomrule
  \end{tabular}
      \caption{Datasets, tasks, details, and number of training examples.}
      \vspace{-0.6cm}
  \label{tasks}
\end{table}

\textbf{Sentiment Analysis.} 
We train our model separately on two sentiment analysis datasets: the Stanford Sentiment Treebank (SST)~\citep{Socher2013EMNLP} and the IMDb dataset~\citep{maas-EtAl:2011:ACL-HLT2011}.
Both of these datasets comprise movie reviews and their sentiment.
We use the binary version of each dataset as well as the five-class version of SST.
For training on SST, we use all sub-trees with length greater than $3$.
SST-2 contains roughly $56,400$ reviews after removing ``neutral'' examples. 
SST-5 contains roughly $94,200$ reviews and does include ``neutral'' examples.
IMDb contains $25,000$ multi-sentence reviews, which we truncate to the first $200$ words. $2,500$ reviews are held out for validation. 

\textbf{Question Classification.} 
For question classification, 
we use the small TREC dataset~\citep{voorhees1999trec} dataset of open-domain, fact-based questions divided into broad semantic categories.
We experiment with both the six-class and fifty-class versions of TREC,
which which refer to as TREC-6 and TREC-50, respectively.
We hold out $452$ examples for validation and leave $5,000$ for training.

\textbf{Entailment.}
For entailment, 
we use the Stanford Natural Language Inference Corpus (SNLI)~\citep{bowman2015snli}, 
which has 550,152 training, 
10,000 validation,
and 10,000 testing examples.
Each example consists of a premise, 
a hypothesis, 
and a label specifying whether the premise entails, contradicts, 
or is neutral with respect to the hypothesis.

\textbf{Question Answering.}
The Stanford Question Answering Dataset
(SQuAD)~\citep{rajpurkar2016squad} 
is a large-scale question answering dataset 
with 87,599 training examples,
10,570 development examples,
and a test set that is not released to the public.
Examples consist of question-answer pairs associated with a paragraph from the English Wikipedia.
SQuAD examples assume that the question is answerable and that the answer is contained verbatim somewhere in the paragraph.

\section{Experiments}
\subsection{Machine Translation}
The MT-LSTM trained on MT-Small obtains an uncased,
tokenized BLEU score of $38.5$ on the Multi30k test set from 2016.
The model trained on MT-Medium obtains an uncased, 
tokenized BLEU score of $25.54$ on the IWSLT test set from 2014.
The MT-LSTM trained on MT-Large obtains an uncased, 
tokenized BLEU score of $28.96$ on the WMT 2016 test set.
These results represent strong baseline machine translation models for their respective datasets. 
Note that, while the smallest dataset has the highest BLEU score, 
it is also a much simpler dataset with a restricted domain.

\textbf{Training Details. } 
When training an MT-LSTM,
we used fixed 300-dimensional word vectors.
We used the CommonCrawl-840B GloVe model for English word vectors,
which were completely fixed during training,
so that the MT-LSTM had to learn how to use the pretrained vectors for translation.
The hidden size of the LSTMs in all MT-LSTMs is 300. 
Because all MT-LSTMs are bidirectional,
they output 600-dimensional vectors.
The model was trained with stochastic gradient descent with a learning rate that began at $1$ and decayed by half each epoch after the validation perplexity increased for the first time.
Dropout with ratio $0.2$ was applied to the inputs and outputs of all layers of the encoder and decoder.

\begin{figure}
\begin{subfigure}{0.48\textwidth}
\includegraphics[width=\linewidth]{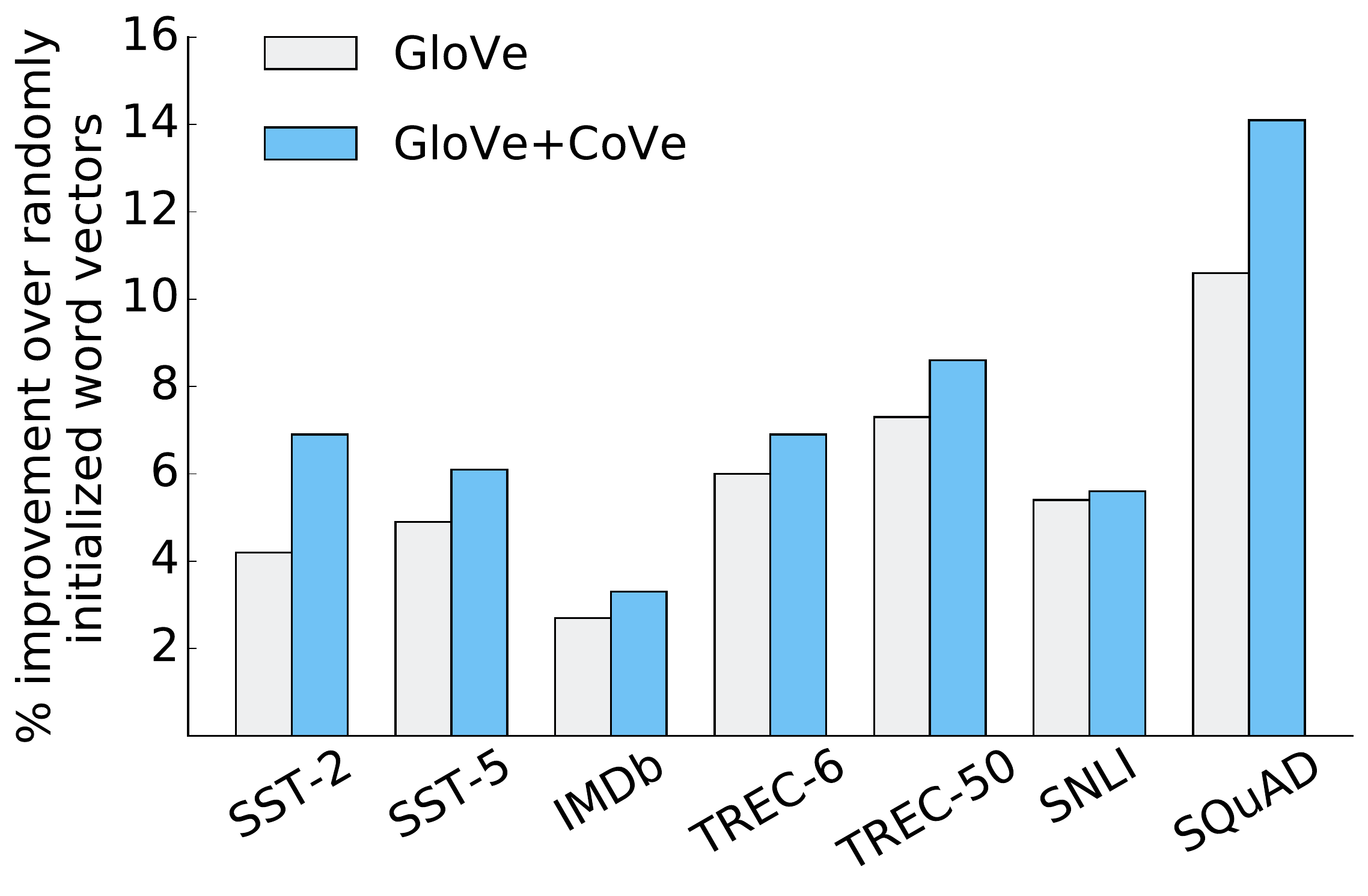}\vspace{-0.2cm}
\caption{CoVe and GloVe} \label{fig:coveGloVe}
\end{subfigure}
\hspace*{\fill} 
\begin{subfigure}{0.48\textwidth}
\includegraphics[width=\linewidth]{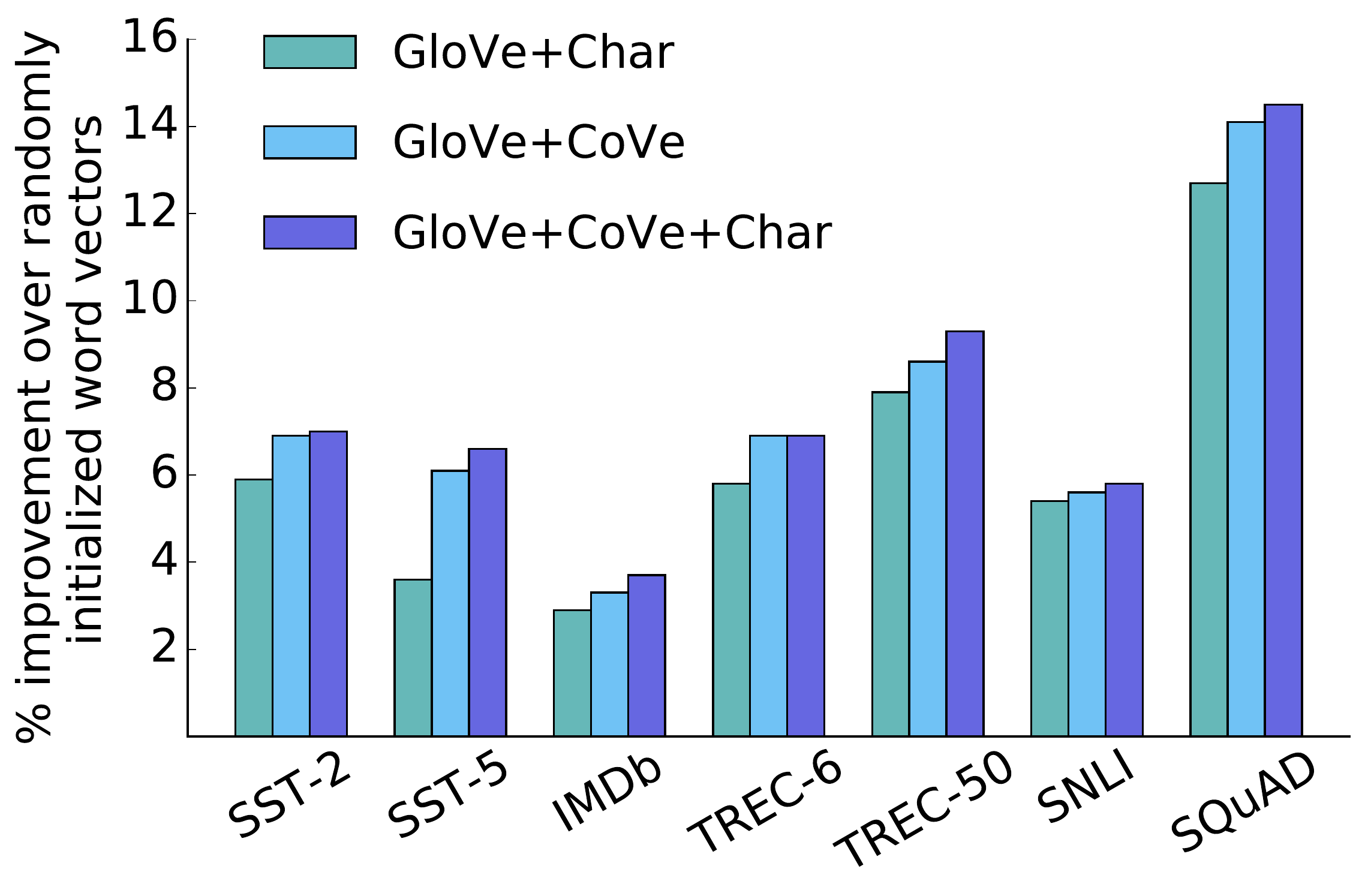}\vspace{-0.2cm}
\caption{CoVe and Characters} \label{fig:coveChar}
\end{subfigure}
\caption{The Benefits of CoVe} \label{fig:bars}
\end{figure}

\begin{table}
\captionsetup{width=.92\textwidth}
  \centering
\begin{tabular}{lccccccc}
\toprule
        &        &       & \multicolumn{5}{c}{GloVe+} \\
\cmidrule(lr){4-8}
Dataset & Random & GloVe & Char & CoVe-S & CoVe-M & CoVe-L & Char+CoVe-L \\
\midrule
SST-2   & 84.2 & 88.4 & 90.1 & 89.0 & 90.9 & 91.1 & \textbf{91.2}\\ 
SST-5   & 48.6 & 53.5 & 52.2 & 54.0 & 54.7 & 54.5 & \textbf{55.2}\\
IMDb    & 88.4 & 91.1 & 91.3 & 90.6 & 91.6 & 91.7 & \textbf{92.1}\\
TREC-6  & 88.9 & 94.9 & 94.7 & 94.7 & 95.1 & 95.8 & \textbf{95.8}\\ 
TREC-50 & 81.9 & 89.2 & 89.8 & 89.6 & 89.6 & 90.5 & \textbf{91.2}\\
SNLI    & 82.3 & 87.7 & 87.7 & 87.3 & 87.5 & 87.9 & \textbf{88.1}\\
SQuAD   & 65.4 & 76.0 & 78.1 & 76.5 & 77.1 & 79.5 & \textbf{79.9}\\
\bottomrule
  \end{tabular}
      \caption{CoVe improves validation performance. CoVe has an advantage over character n-gram embeddings, but using both improves performance further. Models benefit most by using an MT-LSTM trained with MT-Large (CoVe-L). Accuracy is reported for classification tasks, and F1 is reported for SQuAD.}
  \label{validationPerformance}
  \vspace{-0.8cm}
\end{table}
 
\subsection{Classification and Question Answering}
For classification and question answering, 
we explore how varying the input representations affects final performance.
Table~\ref{validationPerformance} contains validation performances for experiments comparing the use of GloVe, character n-grams, CoVe, and combinations of the three.

\textbf{Training Details.}
Unsupervised vectors and MT-LSTMs remain fixed in this set of experiments.
LSTMs have hidden size 300.
Models were trained using Adam with $\alpha=0.001$. 
Dropout was applied before all feedforward layers with dropout ratio $0.1$, $0.2$, or $0.3$.
Maxout networks pool over $4$ channels, reduce dimensionality by $2$, $4$, or $8$, reduce again by $2$, and project to the output dimension.

\textbf{The Benefits of CoVe.}
Figure~\ref{fig:coveGloVe} shows that models that use CoVe alongside GloVe achieve higher validation performance than models that use only GloVe.
Figure~\ref{fig:coveChar} shows that using CoVe in Eq.~\ref{eq:concatVectors} brings larger improvements than using character n-gram embeddings~\citep{Hashimoto2016AJM}. It also shows that altering Eq.~\ref{eq:concatVectors} by additionally appending character n-gram embeddings can boost performance even further for some tasks.
This suggests that the information provided by CoVe is complementary to both the word-level information provided by GloVe as well as the character-level information provided by character n-gram embeddings.

\begin{wrapfigure}[16]{r}{0.62\textwidth}
\vspace{-5.5mm}
  \centering
  \includegraphics[width=\linewidth]{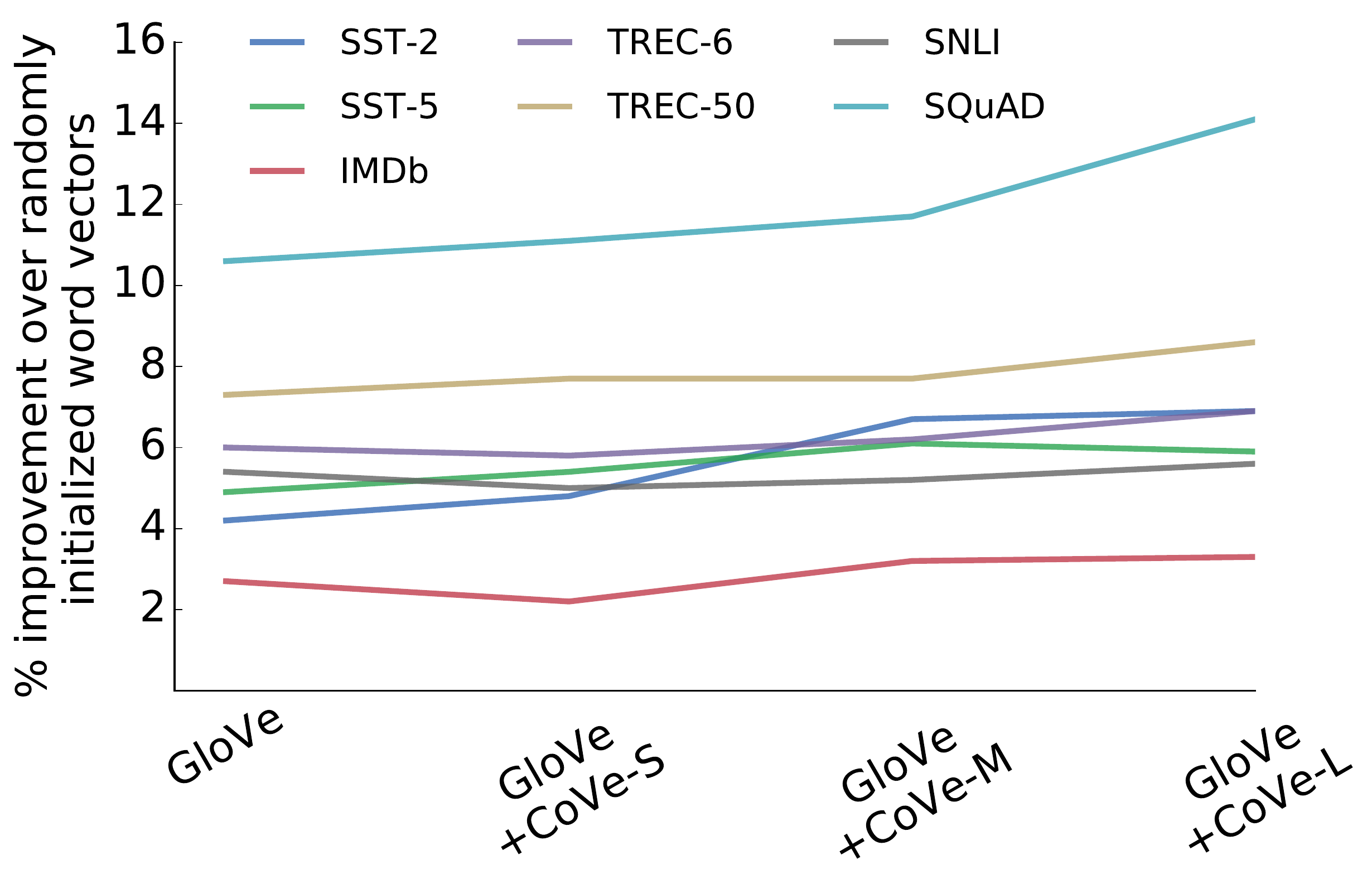}
  \caption{The Effects of MT Training Data
  }
\label{fig:coveComp}
\end{wrapfigure}

\textbf{The Effects of MT Training Data.}
We experimented with different training datasets for the MT-LSTMs to see how varying the MT training data affects the benefits of using CoVe in downstream tasks.
Figure~\ref{fig:coveComp} shows an important trend we can extract from Table~\ref{validationPerformance}. 
There appears to be a positive correlation between the larger MT datasets,
which contain more complex, varied language, 
and the improvement that using CoVe brings to downstream tasks.
This is evidence for our hypothesis that MT data has potential as a large resource for transfer learning in NLP.

\textbf{Test Performance.} 
Table~\ref{testPerf} shows the final test accuracies of our best classification models, 
each of which achieved the highest validation accuracy on its task using GloVe, CoVe, and character n-gram embeddings.
Final test performances on SST-5 and SNLI reached a new state of the art.

\begin{wraptable}{r}{8cm}
    \vspace{-3.5mm}
  \centering
  \begin{tabular}{lcc}
    \toprule
Model & EM  & F1 \\
\midrule
LR~\citep{rajpurkar2016squad} & 40.0 & 51.0 \\
DCR~\citep{yu2017end} & 62.5 & 72.1  \\
hM-LSTM+AP~\citep{wang2017machine} & 64.1 & 73.9 \\
DCN+Char~\citep{Xiong2017} & 65.4 & 75.6 \\
BiDAF~\citep{Seo2017BidirectionalAF} & 68.0 & 77.3 \\
R-NET~\citep{Wang2017} & 71.1 & 79.5 \\
{\it \textbf{DCN+Char+CoVe} } [{\it \textbf{Ours}}] &{\it \textbf{71.3} }&{\it \textbf{79.9} }\\
\bottomrule
  \end{tabular}
      \caption{Exact match and F1 validation scores for single-model question answering.
      }
  \label{squadComps}
    \vspace{-8mm}
\end{wraptable}

Table~\ref{squadComps} shows how the validation exact match and F1 scores of our best SQuAD model compare to the scores of the most recent top models in the literature. 
We did not submit the SQuAD model for testing, 
but the addition of CoVe was enough to push the validation performance of the original DCN,
which already used character n-gram embeddings,
above the validation performance of the published version of the R-NET. 
Test performances are tracked by the SQuAD leaderboard~\footnote{\url{https://rajpurkar.github.io/SQuAD-explorer/}}.

\begin{table}
  \centering
  \setlength\tabcolsep{3.65pt}
\begin{tabular}{llcllc}
    \toprule
 & Model & Test &  & Model & Test\\
\midrule
 \parbox[t]{2mm}{\multirow{6}{*}{\rotatebox[origin=c]{90}{SST-2}}} & P-LSTM~\citep{Wieting2015TowardsUP} & 89.2             & \parbox[t]{2mm}{\multirow{6}{*}{\rotatebox[origin=c]{90}{TREC-6}}} & SVM~\citep{Silva2011FromST} & 95.0           \\
                        & CT-LSTM~\citep{Looks2017DeepLW} & 89.4             &                         & SVM~\citep{Vantu2016QC}  & 95.2              \\
                        & TE-LSTM~\citep{Huang2017EncodingSK} & 89.6             &                         & DSCNN-P~\citep{Zhang2016DependencySC} & 95.6             \\
                        & NSE~\citep{Munkhdalai2016NeuralSE} & 89.7             &                         & {\it BCN+Char+CoVe [Ours] }& {\it95.8 }          \\
                        & {\it BCN+Char+CoVe [Ours]           }          & {\it90.3}          &                         &  TBCNN~\citep{Mou2015DiscriminativeNS}& 96.0             \\
                        & \textbf{bmLSTM~\citep{Radford2017LearningTG}} & \textbf{91.8}             &                         & \textbf{LSTM-CNN~\citep{Zhou2016TextCI}}            & \textbf{96.1} \\
\midrule
 \parbox[t]{2mm}{\multirow{6}{*}{\rotatebox[origin=c]{90}{SST-5}}} & MVN~\citep{Guo2017EndtoEndMN} & 51.5          & \parbox[t]{2mm}{\multirow{6}{*}{\rotatebox[origin=c]{90}{TREC-50}}}& SVM~\citep{Loni2011QuestionCB}&89.0           \\
                        & DMN~\citep{Kumar2016} & 52.1             &                         & SNoW~\citep{Li2006LearningQC} & 89.3             \\
                        & LSTM-CNN~\citep{Zhou2016TextCI} & 52.4             &                         & {\it BCN+Char+CoVe [Ours]        }           & {\it90.2}             \\
                        & TE-LSTM~\citep{Huang2017EncodingSK} & 52.6             &                         & RulesUHC~\citep{Silva2011FromST}       & 90.8             \\
                        & NTI~\citep{Munkhdalai2016NeuralTI} & 53.1             &                         &  SVM~\citep{Vantu2016QC}                    &     91.6      \\
                        & {\it \textbf{BCN+Char+CoVe [Ours]} }           & {\it \textbf{53.7}} &                         &  \textbf{Rules~\citep{Madabushi2016HighAR}}                   & \textbf{97.2}             \\
\midrule
 \parbox[t]{2mm}{\multirow{6}{*}{\rotatebox[origin=c]{90}{IMDb}}}  & {\it BCN+Char+CoVe [Ours]}                     &{\it91.8}          & \parbox[t]{2mm}{\multirow{6}{*}{\rotatebox[origin=c]{90}{SNLI}}}   &DecAtt+Intra~\citep{Parikh2016ADA} &86.8             \\
                        & SA-LSTM~\citep{Dai2015SemisupervisedSL} &92.8             &                         & NTI~\citep{Munkhdalai2016NeuralTI} & 87.3             \\
                        & bmLSTM~\citep{Radford2017LearningTG} &92.9             &                        & re-read LSTM~\citep{Sha2016ReadingAT} & 87.5             \\
                        & TRNN~\citep{Dieng2016TopicRNNAR} &93.8             &                        & btree-LSTM~\citep{Paria2016ANA} & 87.6            \\
               & oh-LSTM~\citep{Johnson2016SupervisedAS} &94.1             &                         & 600D ESIM~\citep{Chen2016EnhancingAC} &  88.0             \\
                        & {\bf Virtual~\citep{Miyato2017AdversarialTM}} &{\bf94.1}            &                         & {\it \textbf{BCN+Char+CoVe [Ours]}}            & {\it\textbf{88.1}} \\
\bottomrule
  \end{tabular}
      \caption{Single model test accuracies for classification tasks.}
  \label{testPerf}
  \vspace{-4mm}
\end{table}

\begin{wraptable}{r}{6cm}
  \vspace{-3.5mm}
\captionsetup{width=5.6cm}
  \centering
\begin{tabular}{lcc}
\toprule
        & \multicolumn{2}{c}{GloVe+Char+} \\
\cmidrule(lr){2-3}
Dataset & Skip-Thought & CoVe-L \\
\midrule
SST-2   & 88.7 & \textbf{91.2}\\ 
SST-5   & 52.1 & \textbf{55.2}\\
TREC-6  & 94.2 & \textbf{95.8}\\ 
TREC-50 & 89.6 & \textbf{91.2}\\
SNLI    & 86.0 &  \textbf{88.1}\\
\bottomrule
  \end{tabular}
      \caption{Classification validation accuracies with skip-thought and CoVe.}
  \label{SkipTable}
\end{wraptable}

\textbf{Comparison to Skip-Thought Vectors.}
~\citet{Kiros2015SkipThoughtV} show how to encode a sentence into a single skip-thought vector that transfers well to a variety of tasks.
Both skip-thought and CoVe pretrain encoders to capture information at a higher level than words. However, skip-thought encoders are trained with an unsupervised method that relies on the final output of the encoder. MT-LSTMs are trained with a supervised method that instead relies on intermediate outputs associated with each input word. Additionally, the $4800$ dimensional skip-thought vectors make training more unstable than using the $600$ dimensional CoVe. Table~\ref{SkipTable} shows that these differences make CoVe more suitable for transfer learning in our classification experiments.

\section{Conclusion}
We introduce an approach for transferring knowledge from an encoder pretrained on machine translation to a variety of downstream NLP tasks. 
In all cases, models that used CoVe from our best, pretrained MT-LSTM performed better than baselines that used random word vector initialization, baselines that used pretrained word vectors from a GloVe model, and baselines that used word vectors from a GloVe model together with character n-gram embeddings.
We hope this is a step towards the goal of building unified NLP models that rely on increasingly more general reusable weights.

The PyTorch code at \url{https://github.com/salesforce/cove} includes an example of how to generate CoVe from the MT-LSTM we used in all of our best models. We hope that making our best MT-LSTM available will encourage further research into shared representations for NLP models.

\small
\bibliography{references}

\begin{thebibliography}{71}
\providecommand{\natexlab}[1]{#1}
\providecommand{\url}[1]{\texttt{#1}}
\expandafter\ifx\csname urlstyle\endcsname\relax
  \providecommand{\doi}[1]{doi: #1}\else
  \providecommand{\doi}{doi: \begingroup \urlstyle{rm}\Url}\fi

\bibitem[Agirre et~al.(2014)Agirre, Banea, Cardie, Cer, Diab, Gonzalez-Agirre,
  Guo, Mihalcea, Rigau, and Wiebe]{Agirre2014SemEval2014T1}
E.~Agirre, C.~Banea, C.~Cardie, D.~M. Cer, M.~T. Diab, A.~Gonzalez-Agirre,
  W.~Guo, R.~Mihalcea, G.~Rigau, and J.~Wiebe.
\newblock {SemEval-2014 Task} 10: Multilingual semantic textual similarity.
\newblock In \emph{SemEval@COLING}, 2014.

\bibitem[Bahdanau et~al.(2015)Bahdanau, Cho, and Bengio]{Bahdanau2015}
D.~Bahdanau, K.~Cho, and Y.~Bengio.
\newblock Neural machine translation by jointly learning to align and
  translate.
\newblock In \emph{ICLR}, 2015.

\bibitem[Bowman et~al.(2014)Bowman, Potts, and Manning]{Bowman2014}
S.~R. Bowman, C.~Potts, and C.~D. Manning.
\newblock Recursive neural networks for learning logical semantics.
\newblock \emph{CoRR}, abs/1406.1827, 2014.

\bibitem[Bowman et~al.(2015)Bowman, Angeli, Potts, and Manning]{bowman2015snli}
S.~R. Bowman, G.~Angeli, C.~Potts, and C.~D. Manning.
\newblock A large annotated corpus for learning natural language inference.
\newblock In \emph{Proceedings of the 2015 Conference on Empirical Methods in
  Natural Language Processing (EMNLP)}. Association for Computational
  Linguistics, 2015.

\bibitem[Cettolo et~al.(2015)Cettolo, Niehues, St{\"u}ker, Bentivogli, Cattoni,
  and Federico]{cettolo2015iwslt}
M.~Cettolo, J.~Niehues, S.~St{\"u}ker, L.~Bentivogli, R.~Cattoni, and
  M.~Federico.
\newblock The {IWSLT} 2015 evaluation campaign.
\newblock In \emph{IWSLT}, 2015.

\bibitem[Chen et~al.(2016)Chen, Zhu, Ling, Wei, and Jiang]{Chen2016EnhancingAC}
Q.~Chen, X.-D. Zhu, Z.-H. Ling, S.~Wei, and H.~Jiang.
\newblock Enhancing and combining sequential and tree {LSTM} for natural
  language inference.
\newblock \emph{CoRR}, abs/1609.06038, 2016.

\bibitem[Collobert et~al.(2011)Collobert, Weston, Bottou, Karlen, Kavukcuoglu,
  and Kuksa]{Collobert2011}
R.~Collobert, J.~Weston, L.~Bottou, M.~Karlen, K.~Kavukcuoglu, and P.~Kuksa.
\newblock Natural language processing (almost) from scratch.
\newblock \emph{JMLR}, 12:\penalty0 2493--2537, 2011.

\bibitem[Conneau et~al.(2017)Conneau, Kiela, Schwenk, Barrault, and
  Bordes]{conneau2017supervised}
A.~Conneau, D.~Kiela, H.~Schwenk, L.~Barrault, and A.~Bordes.
\newblock Supervised learning of universal sentence representations from
  natural language inference data.
\newblock \emph{arXiv preprint arXiv:1705.02364}, 2017.

\bibitem[da~Silva et~al.(2011)da~Silva, Coheur, Mendes, and
  Wichert]{Silva2011FromST}
J.~P. C.~G. da~Silva, L.~Coheur, A.~C. Mendes, and A.~Wichert.
\newblock From symbolic to sub-symbolic information in question classification.
\newblock \emph{Artif. Intell. Rev.}, 35:\penalty0 137--154, 2011.

\bibitem[Dai and Le(2015)]{Dai2015SemisupervisedSL}
A.~M. Dai and Q.~V. Le.
\newblock Semi-supervised sequence learning.
\newblock In \emph{NIPS}, 2015.

\bibitem[Deng et~al.(2009)Deng, Dong, Socher, Li, Li, and
  Fei-Fei]{Deng2009ImageNetAL}
J.~Deng, W.~Dong, R.~Socher, L.-J. Li, K.~Li, and L.~Fei-Fei.
\newblock {ImageNet}: A large-scale hierarchical image database.
\newblock \emph{2009 IEEE Conference on Computer Vision and Pattern
  Recognition}, pages 248--255, 2009.

\bibitem[Dieng et~al.(2016)Dieng, Wang, Gao, and Paisley]{Dieng2016TopicRNNAR}
A.~B. Dieng, C.~Wang, J.~Gao, and J.~W. Paisley.
\newblock {TopicRNN}: A recurrent neural network with long-range semantic
  dependency.
\newblock \emph{CoRR}, abs/1611.01702, 2016.

\bibitem[Dong and Lapata(2016)]{Dong2016LanguageTL}
L.~Dong and M.~Lapata.
\newblock Language to logical form with neural attention.
\newblock \emph{CoRR}, abs/1601.01280, 2016.

\bibitem[Fukui et~al.(2016)Fukui, Park, Yang, Rohrbach, Darrell, and
  Rohrbach]{fukui2016multimodal}
A.~Fukui, D.~H. Park, D.~Yang, A.~Rohrbach, T.~Darrell, and M.~Rohrbach.
\newblock Multimodal compact bilinear pooling for visual question answering and
  visual grounding.
\newblock \emph{arXiv preprint arXiv:1606.01847}, 2016.

\bibitem[Girshick et~al.(2014)Girshick, Donahue, Darrell, and
  Malik]{girshick2014rich}
R.~Girshick, J.~Donahue, T.~Darrell, and J.~Malik.
\newblock Rich feature hierarchies for accurate object detection and semantic
  segmentation.
\newblock In \emph{Proceedings of the IEEE conference on computer vision and
  pattern recognition}, pages 580--587, 2014.

\bibitem[Goodfellow et~al.(2013)Goodfellow, Warde-Farley, Mirza, Courville, and
  Bengio]{Goodfellow2013MaxoutN}
I.~J. Goodfellow, D.~Warde-Farley, M.~Mirza, A.~C. Courville, and Y.~Bengio.
\newblock Maxout networks.
\newblock In \emph{ICML}, 2013.

\bibitem[Graves and Schmidhuber(2005)]{graves2005framewise}
A.~Graves and J.~Schmidhuber.
\newblock Framewise phoneme classification with bidirectional {LSTM} and other
  neural network architectures.
\newblock \emph{Neural Networks}, 18\penalty0 (5):\penalty0 602--610, 2005.

\bibitem[Guo et~al.(2017)Guo, Cherry, and Su]{Guo2017EndtoEndMN}
H.~Guo, C.~Cherry, and J.~Su.
\newblock End-to-end multi-view networks for text classification.
\newblock \emph{CoRR}, abs/1704.05907, 2017.

\bibitem[Hashimoto et~al.(2016)Hashimoto, Xiong, Tsuruoka, and
  Socher]{Hashimoto2016AJM}
K.~Hashimoto, C.~Xiong, Y.~Tsuruoka, and R.~Socher.
\newblock A joint many-task model: Growing a neural network for multiple {NLP}
  tasks.
\newblock \emph{CoRR}, abs/1611.01587, 2016.

\bibitem[He et~al.(2016)He, Zhang, Ren, and Sun]{he2016deep}
K.~He, X.~Zhang, S.~Ren, and J.~Sun.
\newblock Deep residual learning for image recognition.
\newblock In \emph{Proceedings of the IEEE Conference on Computer Vision and
  Pattern Recognition}, pages 770--778, 2016.

\bibitem[Hill et~al.(2016)Hill, Cho, and Korhonen]{Hill2016LearningDR}
F.~Hill, K.~Cho, and A.~Korhonen.
\newblock Learning distributed representations of sentences from unlabelled
  data.
\newblock In \emph{HLT-NAACL}, 2016.

\bibitem[Hill et~al.(2017)Hill, Cho, Jean, and Bengio]{Hill2017}
F.~Hill, K.~Cho, S.~Jean, and Y.~Bengio.
\newblock The representational geometry of word meanings acquired by neural
  machine translation models.
\newblock \emph{Machine Translation}, pages 1--16, 2017.
\newblock ISSN 1573-0573.
\newblock \doi{10.1007/s10590-017-9194-2}.
\newblock URL \url{http://dx.doi.org/10.1007/s10590-017-9194-2}.

\bibitem[Huang et~al.(2017)Huang, Qian, and Zhu]{Huang2017EncodingSK}
M.~Huang, Q.~Qian, and X.~Zhu.
\newblock Encoding syntactic knowledge in neural networks for sentiment
  classification.
\newblock \emph{ACM Trans. Inf. Syst.}, 35:\penalty0 26:1--26:27, 2017.

\bibitem[Ioffe and Szegedy(2015)]{Ioffe2015BatchNA}
S.~Ioffe and C.~Szegedy.
\newblock Batch normalization: Accelerating deep network training by reducing
  internal covariate shift.
\newblock In \emph{ICML}, 2015.

\bibitem[Johnson and Zhang(2016)]{Johnson2016SupervisedAS}
R.~Johnson and T.~Zhang.
\newblock Supervised and semi-supervised text categorization using {LSTM} for
  region embeddings.
\newblock In \emph{ICML}, 2016.

\bibitem[Kiros et~al.(2015)Kiros, Zhu, Salakhutdinov, Zemel, Urtasun, Torralba,
  and Fidler]{Kiros2015SkipThoughtV}
R.~Kiros, Y.~Zhu, R.~Salakhutdinov, R.~S. Zemel, R.~Urtasun, A.~Torralba, and
  S.~Fidler.
\newblock Skip-thought vectors.
\newblock In \emph{NIPS}, 2015.

\bibitem[{Klein} et~al.(2017){Klein}, {Kim}, {Deng}, {Senellart}, and
  {Rush}]{2017opennmt}
G.~{Klein}, Y.~{Kim}, Y.~{Deng}, J.~{Senellart}, and A.~M. {Rush}.
\newblock {OpenNMT}: Open-source toolkit for neural machine translation.
\newblock \emph{ArXiv e-prints}, 2017.

\bibitem[Koehn et~al.(2007)Koehn, Hoang, Birch, Callison-Burch, Federico,
  Bertoldi, Cowan, Shen, Moran, Zens, Dyer, Bojar, Constantin, and
  Herbst]{Koehn2007}
P.~Koehn, H.~Hoang, A.~Birch, C.~Callison-Burch, M.~Federico, N.~Bertoldi,
  B.~Cowan, W.~Shen, C.~Moran, R.~Zens, C.~Dyer, O.~Bojar, A.~Constantin, and
  E.~Herbst.
\newblock Moses: Open source toolkit for statistical machine translation.
\newblock In \emph{ACL}, 2007.

\bibitem[Krizhevsky et~al.(2012)Krizhevsky, Sutskever, and
  Hinton]{krizhevsky2012imagenet}
A.~Krizhevsky, I.~Sutskever, and G.~E. Hinton.
\newblock Imagenet classification with deep convolutional neural networks.
\newblock In \emph{Advances in neural information processing systems}, pages
  1097--1105, 2012.

\bibitem[Kumar et~al.(2016)Kumar, Irsoy, Ondruska, Iyyer, Bradbury, Gulrajani,
  Zhong, Paulus, and Socher]{Kumar2016}
A.~Kumar, O.~Irsoy, P.~Ondruska, M.~Iyyer, J.~Bradbury, I.~Gulrajani, V.~Zhong,
  R.~Paulus, and R.~Socher.
\newblock Ask me anything: Dynamic memory networks for natural language
  processing.
\newblock In \emph{ICML}, 2016.

\bibitem[Li and Roth(2006)]{Li2006LearningQC}
X.~Li and D.~Roth.
\newblock Learning question classifiers: The role of semantic information.
\newblock \emph{Natural Language Engineering}, 12:\penalty0 229--249, 2006.

\bibitem[Loni et~al.(2011)Loni, van Tulder, Wiggers, Tax, and
  Loog]{Loni2011QuestionCB}
B.~Loni, G.~van Tulder, P.~Wiggers, D.~M.~J. Tax, and M.~Loog.
\newblock Question classification by weighted combination of lexical, syntactic
  and semantic features.
\newblock In \emph{TSD}, 2011.

\bibitem[Looks et~al.(2017)Looks, Herreshoff, Hutchins, and
  Norvig]{Looks2017DeepLW}
M.~Looks, M.~Herreshoff, D.~Hutchins, and P.~Norvig.
\newblock Deep learning with dynamic computation graphs.
\newblock \emph{CoRR}, abs/1702.02181, 2017.

\bibitem[Lu et~al.(2016)Lu, Xiong, Parikh, and Socher]{lu2016knowing}
J.~Lu, C.~Xiong, D.~Parikh, and R.~Socher.
\newblock Knowing when to look: Adaptive attention via a visual sentinel for
  image captioning.
\newblock \emph{arXiv preprint arXiv:1612.01887}, 2016.

\bibitem[Luong et~al.(2015)Luong, Pham, and Manning]{Luong2015EffectiveAT}
T.~Luong, H.~Pham, and C.~D. Manning.
\newblock Effective approaches to attention-based neural machine translation.
\newblock In \emph{EMNLP}, 2015.

\bibitem[Maas et~al.(2011)Maas, Daly, Pham, Huang, Ng, and
  Potts]{maas-EtAl:2011:ACL-HLT2011}
A.~L. Maas, R.~E. Daly, P.~T. Pham, D.~Huang, A.~Y. Ng, and C.~Potts.
\newblock Learning word vectors for sentiment analysis.
\newblock In \emph{Proceedings of the 49th Annual Meeting of the Association
  for Computational Linguistics: Human Language Technologies}, pages 142--150,
  Portland, Oregon, USA, June 2011. Association for Computational Linguistics.
\newblock URL \url{http://www.aclweb.org/anthology/P11-1015}.

\bibitem[Madabushi and Lee(2016)]{Madabushi2016HighAR}
H.~T. Madabushi and M.~Lee.
\newblock High accuracy rule-based question classification using question
  syntax and semantics.
\newblock In \emph{COLING}, 2016.

\bibitem[Mikolov et~al.(2013)Mikolov, Chen, Corrado, and Dean]{Mikolov2013b}
T.~Mikolov, K.~Chen, G.~Corrado, and J.~Dean.
\newblock Efficient estimation of word representations in vector space.
\newblock In \emph{ICLR (workshop)}, 2013.

\bibitem[Min et~al.(2017)Min, Seo, and Hajishirzi]{min2017question}
S.~Min, M.~Seo, and H.~Hajishirzi.
\newblock Question answering through transfer learning from large fine-grained
  supervision data.
\newblock 2017.

\bibitem[Miyato et~al.(2017)Miyato, Dai, and
  Goodfellow]{Miyato2017AdversarialTM}
T.~Miyato, A.~M. Dai, and I.~Goodfellow.
\newblock Adversarial training methods for semi-supervised text classification.
\newblock 2017.

\bibitem[Mou et~al.(2015)Mou, Peng, Li, Xu, Zhang, and
  Jin]{Mou2015DiscriminativeNS}
L.~Mou, H.~Peng, G.~Li, Y.~Xu, L.~Zhang, and Z.~Jin.
\newblock Discriminative neural sentence modeling by tree-based convolution.
\newblock In \emph{EMNLP}, 2015.

\bibitem[Munkhdalai and Yu(2016{\natexlab{a}})]{Munkhdalai2016NeuralSE}
T.~Munkhdalai and H.~Yu.
\newblock Neural semantic encoders.
\newblock \emph{CoRR}, abs/1607.04315, 2016{\natexlab{a}}.

\bibitem[Munkhdalai and Yu(2016{\natexlab{b}})]{Munkhdalai2016NeuralTI}
T.~Munkhdalai and H.~Yu.
\newblock Neural tree indexers for text understanding.
\newblock \emph{CoRR}, abs/1607.04492, 2016{\natexlab{b}}.

\bibitem[Nair and Hinton(2010)]{Nair10}
V.~Nair and G.~E. Hinton.
\newblock Rectified linear units improve restricted {Boltzmann} machines.
\newblock In \emph{ICML}, 2010.

\bibitem[Nallapati et~al.(2016)Nallapati, Zhou, dos Santos, Çaglar
  G\"{u}lçehre, and Xiang]{Nallapati2016AbstractiveTS}
R.~Nallapati, B.~Zhou, C.~N. dos Santos, Çaglar G\"{u}lçehre, and B.~Xiang.
\newblock Abstractive text summarization using sequence-to-sequence {RNNs} and
  beyond.
\newblock In \emph{CoNLL}, 2016.

\bibitem[Paria et~al.(2016)Paria, Annervaz, Dukkipati, Chatterjee, and
  Podder]{Paria2016ANA}
B.~Paria, K.~M. Annervaz, A.~Dukkipati, A.~Chatterjee, and S.~Podder.
\newblock A neural architecture mimicking humans end-to-end for natural
  language inference.
\newblock \emph{CoRR}, abs/1611.04741, 2016.

\bibitem[Parikh et~al.(2016)Parikh, Tackstrom, Das, and
  Uszkoreit]{Parikh2016ADA}
A.~P. Parikh, O.~Tackstrom, D.~Das, and J.~Uszkoreit.
\newblock A decomposable attention model for natural language inference.
\newblock In \emph{EMNLP}, 2016.

\bibitem[Pennington et~al.(2014)Pennington, Socher, and
  Manning]{Pennington2014}
J.~Pennington, R.~Socher, and C.~D. Manning.
\newblock Glove: Global vectors for word representation.
\newblock In \emph{EMNLP}, 2014.

\bibitem[Qi et~al.(2016)Qi, Zhang, Qin, Yao, Huang, Lim, and
  Yang]{qi2016hedged}
Y.~Qi, S.~Zhang, L.~Qin, H.~Yao, Q.~Huang, J.~Lim, and M.-H. Yang.
\newblock Hedged deep tracking.
\newblock In \emph{Proceedings of the IEEE Conference on Computer Vision and
  Pattern Recognition}, pages 4303--4311, 2016.

\bibitem[Radford et~al.(2017)Radford, J{\'o}zefowicz, and
  Sutskever]{Radford2017LearningTG}
A.~Radford, R.~J{\'o}zefowicz, and I.~Sutskever.
\newblock Learning to generate reviews and discovering sentiment.
\newblock \emph{CoRR}, abs/1704.01444, 2017.

\bibitem[Rajpurkar et~al.(2016)Rajpurkar, Zhang, Lopyrev, and
  Liang]{rajpurkar2016squad}
P.~Rajpurkar, J.~Zhang, K.~Lopyrev, and P.~Liang.
\newblock {SQuAD}: 100,000+ questions for machine comprehension of text.
\newblock \emph{arXiv preprint arXiv:1606.05250}, 2016.

\bibitem[Ramachandran et~al.(2016)Ramachandran, Liu, and
  Le]{Ramachandran2016UnsupervisedPF}
P.~Ramachandran, P.~J. Liu, and Q.~V. Le.
\newblock Unsupervised pretraining for sequence to sequence learning.
\newblock \emph{CoRR}, abs/1611.02683, 2016.

\bibitem[Saenko et~al.(2010)Saenko, Kulis, Fritz, and
  Darrell]{Saenko2010AdaptingVC}
K.~Saenko, B.~Kulis, M.~Fritz, and T.~Darrell.
\newblock Adapting visual category models to new domains.
\newblock In \emph{ECCV}, 2010.

\bibitem[Seo et~al.(2017)Seo, Kembhavi, Farhadi, and
  Hajishirzi]{Seo2017BidirectionalAF}
M.~Seo, A.~Kembhavi, A.~Farhadi, and H.~Hajishirzi.
\newblock Bidirectional attention flow for machine comprehension.
\newblock \emph{ICLR}, 2017.

\bibitem[Sha et~al.(2016)Sha, Chang, Sui, and Li]{Sha2016ReadingAT}
L.~Sha, B.~Chang, Z.~Sui, and S.~Li.
\newblock Reading and thinking: Re-read {LSTM} unit for textual entailment
  recognition.
\newblock In \emph{COLING}, 2016.

\bibitem[Simonyan and Zisserman(2014)]{simonyan2014very}
K.~Simonyan and A.~Zisserman.
\newblock Very deep convolutional networks for large-scale image recognition.
\newblock \emph{arXiv preprint arXiv:1409.1556}, 2014.

\bibitem[Socher et~al.(2013)Socher, Perelygin, Wu, Chuang, Manning, Ng, and
  Potts]{Socher2013EMNLP}
R.~Socher, A.~Perelygin, J.~Wu, J.~Chuang, C.~Manning, A.~Ng, and C.~Potts.
\newblock Recursive deep models for semantic compositionality over a sentiment
  treebank.
\newblock In \emph{EMNLP}, 2013.

\bibitem[Socher et~al.(2014)Socher, Karpathy, Le, Manning, and
  Ng]{Socher2014TACL}
R.~Socher, A.~Karpathy, Q.~V. Le, C.~D. Manning, and A.~Y. Ng.
\newblock Grounded compositional semantics for finding and describing images
  with sentences.
\newblock In \emph{ACL}, 2014.

\bibitem[Specia et~al.(2016)Specia, Frank, Sima’an, and Elliott]{Specia2016}
L.~Specia, S.~Frank, K.~Sima’an, and D.~Elliott.
\newblock A shared task on multimodal machine translation and crosslingual
  image description.
\newblock In \emph{WMT}, 2016.

\bibitem[Sutskever et~al.(2014)Sutskever, Vinyals, and Le]{Sutskever2014}
I.~Sutskever, O.~Vinyals, and Q.~V. Le.
\newblock Sequence to sequence learning with neural networks.
\newblock In \emph{NIPS}, 2014.

\bibitem[Van-Tu and Anh-Cuong(2016)]{Vantu2016QC}
N.~Van-Tu and L.~Anh-Cuong.
\newblock Improving question classification by feature extraction and
  selection.
\newblock \emph{Indian Journal of Science and Technology}, 9\penalty0 (17),
  2016.

\bibitem[Voorhees and Tice(1999)]{voorhees1999trec}
E.~M. Voorhees and D.~M. Tice.
\newblock The {TREC-8} question answering track evaluation.
\newblock In \emph{TREC}, volume 1999, page~82, 1999.

\bibitem[Wang and Jiang(2017)]{wang2017machine}
S.~Wang and J.~Jiang.
\newblock Machine comprehension using {Match-LSTM} and answer pointer.
\newblock 2017.

\bibitem[Wang et~al.(2017)Wang, Yang, Wei, Chang, and Zhou]{Wang2017}
W.~Wang, N.~Yang, F.~Wei, B.~Chang, and M.~Zhou.
\newblock Gated self-matching networks for reading comprehension and question
  answering.
\newblock 2017.

\bibitem[Wieting et~al.(2016)Wieting, Bansal, Gimpel, and
  Livescu]{Wieting2015TowardsUP}
J.~Wieting, M.~Bansal, K.~Gimpel, and K.~Livescu.
\newblock Towards universal paraphrastic sentence embeddings.
\newblock In \emph{ICLR}, 2016.

\bibitem[Xiong et~al.(2016)Xiong, Merity, and Socher]{xiong2016dynamic}
C.~Xiong, S.~Merity, and R.~Socher.
\newblock Dynamic memory networks for visual and textual question answering.
\newblock In \emph{Proceedings of The 33rd International Conference on Machine
  Learning}, pages 2397--2406, 2016.

\bibitem[Xiong et~al.(2017)Xiong, Zhong, and Socher]{Xiong2017}
C.~Xiong, V.~Zhong, and R.~Socher.
\newblock Dynamic coattention networks for question answering.
\newblock \emph{ICRL}, 2017.

\bibitem[Yu et~al.(2017)Yu, Zhang, Hasan, Yu, Xiang, and Zhou]{yu2017end}
Y.~Yu, W.~Zhang, K.~Hasan, M.~Yu, B.~Xiang, and B.~Zhou.
\newblock End-to-end reading comprehension with dynamic answer chunk ranking.
\newblock \emph{ICLR}, 2017.

\bibitem[Zhang et~al.(2016)Zhang, Lee, and Radev]{Zhang2016DependencySC}
R.~Zhang, H.~Lee, and D.~R. Radev.
\newblock Dependency sensitive convolutional neural networks for modeling
  sentences and documents.
\newblock In \emph{HLT-NAACL}, 2016.

\bibitem[Zhou et~al.(2016)Zhou, Qi, Zheng, Xu, Bao, and Xu]{Zhou2016TextCI}
P.~Zhou, Z.~Qi, S.~Zheng, J.~Xu, H.~Bao, and B.~Xu.
\newblock Text classification improved by integrating bidirectional {LSTM} with
  two-dimensional max pooling.
\newblock In \emph{COLING}, 2016.

\bibitem[Zhu et~al.(2011)Zhu, Chen, Lu, Pan, Xue, Yu, and
  Yang]{Zhu2011HeterogeneousTL}
Y.~Zhu, Y.~Chen, Z.~Lu, S.~J. Pan, G.-R. Xue, Y.~Yu, and Q.~Yang.
\newblock Heterogeneous transfer learning for image classification.
\newblock In \emph{AAAI}, 2011.

\end{thebibliography}
\bibliographystyle{abbrvnat}

\end{document}